\documentclass[10pt,twocolumn,letterpaper]{article}

\usepackage{cvpr}              % To produce the CAMERA-READY version
\usepackage{graphicx}
\usepackage{amsmath}
\usepackage{amssymb}
\usepackage{booktabs}

\usepackage[ruled, vlined]{algorithm2e} 
\usepackage{bm}
\usepackage{enumitem}
\usepackage{makecell}
\usepackage{multirow}
\usepackage{tabularx}
\usepackage{xcolor}
\usepackage[normalem]{ulem}

\usepackage[accsupp]{axessibility} % Improves PDF readability for those with disabilities.

\usepackage[pagebackref,breaklinks,colorlinks]{hyperref}

\usepackage[capitalize]{cleveref}
\crefname{section}{Sec.}{Secs.}
\Crefname{section}{Section}{Sections}
\Crefname{table}{Table}{Tables}
\crefname{table}{Tab.}{Tabs.}

\def\cvprPaperID{9} % *** Enter the CVPR Paper ID here
\def\confName{CVPR}
\def\confYear{2022}

\begin{document}

\title{Weakly-Supervised Action Detection Guided by Audio Narration}

\author{Keren Ye\thanks{Work done at University of Pittsburgh}\\
Cruise\\
San Francisco, CA, USA\\
{\tt\small yekeren.cn@gmail.com}
\and
Adriana Kovashka\\
University of Pittsburgh\\
Pittsburgh, PA, USA\\
{\tt\small kovashka@cs.pitt.edu}
}
\maketitle

%%%%%%%%% ABSTRACT
\begin{abstract}
  Videos are more well-organized curated data sources for visual concept learning than images. Unlike the 2-dimensional images which only involve the spatial information, the additional temporal dimension bridges and synchronizes multiple modalities. However, in most video detection benchmarks, these additional modalities are not fully utilized. For example, EPIC Kitchens is the largest dataset in first-person (egocentric) vision, yet it still relies on crowdsourced information to refine the action boundaries to provide instance-level action annotations.
  
  We explored how to eliminate the expensive annotations in video detection data which provide refined boundaries. We propose a model to learn from the narration supervision and utilize multimodal features, including RGB, motion flow, and ambient sound. Our model learns to attend to the frames related to the narration label while suppressing the irrelevant frames from being used. Our experiments show that noisy audio narration suffices to learn a good action detection model, thus reducing annotation expenses.

\end{abstract}

%%%%%%%%% BODY TEXT
\section{Introduction}
\label{sec:intro}

Inexpensive and informative side information such as soundtracks and closed captions widely exist in videos. In addition, videos involve a temporal dimension, which synchronizes this side information with the video frames. Both the side information and the synchronization nature provide a good chance for self-learning. However, it is still challenging to achieve the goal of localizing specific actions using self-learning. On the one hand, which side information to use as supervision is unclear. On the other hand, the multiple modalities in videos such as RGB frames, motion features, and ambient sound need to be explored. Due to these complex factors, the idea of adopting the abundant side information to predict instance-level action detection results did not gather enough attention.

\begin{figure}[t]
    \centering
    \includegraphics[width=1.0\linewidth]{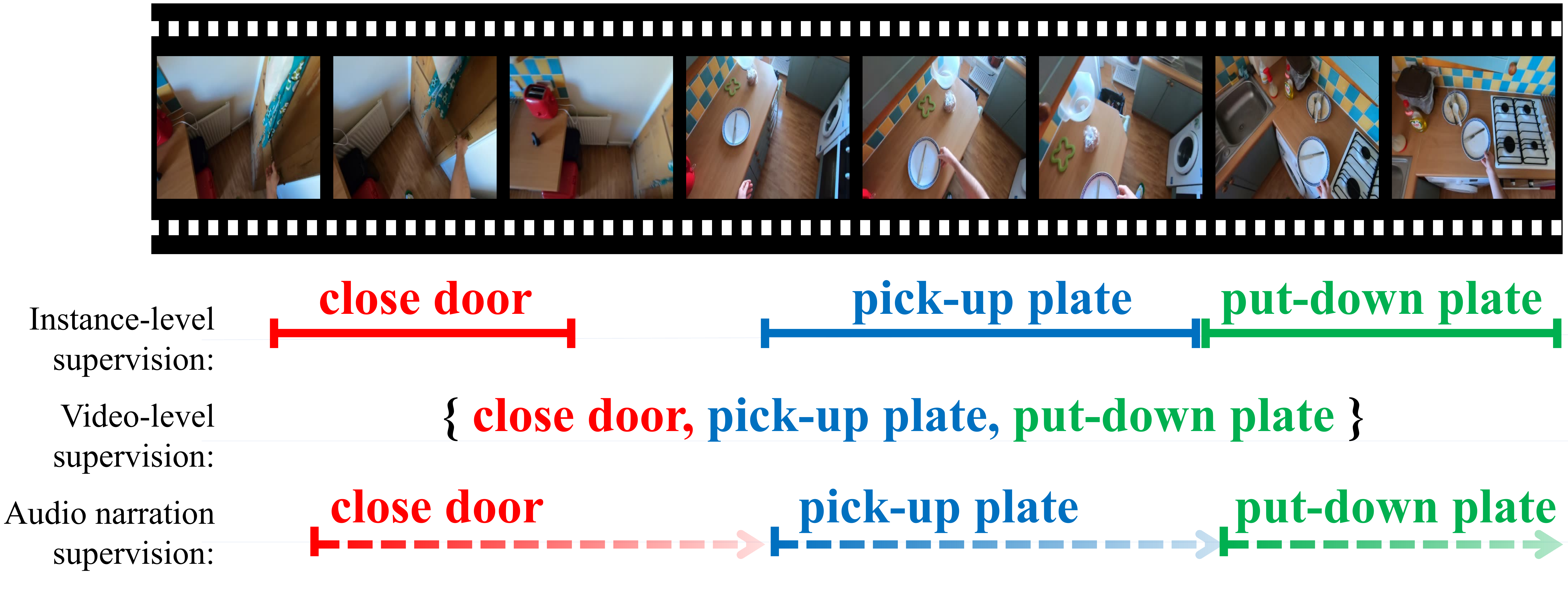}
    \caption[Instance-Level, Video-Level, and Audio Narration Supervisions]{Demonstration of the instance-level, video-level, and audio narration supervisions. The audio narration supervision in the EPIC Kitchens dataset only includes an imprecise start time, while we use this cheap to annotate data to learn a video action detection model.}
    \label{fig:concept}
\end{figure}

This paper will explore audio narrations in the untrimmed video action detection task. As the audio narrations roughly match the video frames, they provide a good signal for localizing actions in the temporal domain. We first distinguish the narration annotations from the instance-level or video-level annotations. 
We show in Fig.~\ref{fig:concept} an example video clip from the EPIC Kitchens dataset~\cite{Damen_2018_ECCV}, as well as the different forms of supervision.
The instance-level annotations are defined by triplets (start time, end time, action class). Models trained under the fully-supervised setting can use this form of supervision to generate temporal action proposals \cite{Lin_2018_ECCV, Lin_2019_ICCV}, or perform action detection \cite{Feichtenhofer_2019_ICCV,Gkioxari_2015_CVPR,Ma_2016_CVPR,Shou_2016_CVPR,Soomro_2015_ICCV,Yeung_2016_CVPR,Zhao_2017_ICCV}. The major benefit of instance-level data is that the resulting models are usually boundary-sensitive, thus the foreground and background are clearly distinguished by the detection scores, resulting in high average precision.

However, fully annotating a video dataset with instance-level labels is time-consuming. Thus, methods \cite{Singh_2017_ICCV,Nguyen_2018_CVPR,Nguyen_2019_ICCV,Paul_2018_ECCV,Shou_2018_ECCV,Wang_2017_CVPR} focus on the weakly supervised action detection (WSAD), which only requires video-level labels. These methods assume the video to be a bag of actions and use multi-label cross-entropy loss to optimize. One disadvantage of WSAD is that it assumes only a few classes per video (e.g., $<$ 5, see Tab.~\ref{tab:datasets}). Hence, it is not applicable in real cases. In an extreme scenario, a 2-hour untrimmed video may consist of all the action classes, thus the video-level label is too coarse %for the model 
to learn a good detector.

This paper explores the audio narration supervision in the EPIC Kitchens~\cite{Damen_2018_ECCV} dataset. In addition to the ambient soundtrack in the dataset, videos in EPIC Kitchens were narrated by the annotators. 
%to gather an extra narration audio track. 
The resulting audio narration track is later transcribed into texts and then parsed into action classes (verb + noun) using a dependency parser, resulting in the forms we use (see Fig.~\ref{fig:concept} bottom). Our goal is to learn action detectors utilizing this form of cheap annotations, and we expect the performance to be comparable to the models learned using instance-level supervision.

\begin{table}[t]
    \centering
    \setlength{\tabcolsep}{3pt}
    \begin{tabularx}{1.0\linewidth}{c|*{3}{>{\centering\arraybackslash}X}}
    \Xhline{2\arrayrulewidth}
        Dataset & Avg. video length (secs) & Avg. classes per video & Avg. actions per video\\
    \Xhline{2\arrayrulewidth}
        THUMOS 14 & 209 & 1.08 & 15.01 \\
        EPIC Kitchens & 477 & 34.87 & 89.36 \\
    \Xhline{2\arrayrulewidth}
    \end{tabularx}
    \caption[Datasets Information]{Datasets information. Most WSAD methods use THUMOS 14~\cite{idrees2017thumos}, in which there is only 1 action class per video. We explore single-timestamp audio narration annotations in EPIC Kichens~\cite{Heilbron_2015_CVPR}. }
    \label{tab:datasets}
\end{table}

The major challenge of using audio narrations is that the \textbf{\textit{annotations are noisy}}. Unlike the instance-level annotation: (1) narration annotations' start timestamps are not precise in that they may overlap with the previous action, and (2) the end time is unknown since the narrators provide no hints about it. One can assume the end time to be before the start time of the following action, but the frames in between are a gray area, and their membership is uncertain. Therefore, narrations provide a trade-off between accurate instance-level annotation and cheap and fast video-level annotation. One needs to model the uncertainty to use them.

To use the narrations to learn the action detection model, we first cut the untrimmed videos into clips using the single timestamp (start time) of the audio narration annotations (see Fig.~\ref{fig:concept}). Thus, each clip can be treated as a mixture of actions, given that the boundaries are imprecise. Then, the association between the frames in the clip and the clip-level action class could be solved as a Multiple Instance Learning (MIL) problem. Compared to the common WSAD methods such as ~\cite{Nguyen_2018_CVPR,Nguyen_2019_ICCV} which only distinguish between foreground and background (see statistics of THUMOS 14 in Tab.~\ref{tab:datasets}), \textbf{the background in our clip may involve other semantically meaningful actions}. So, we design a class-aware attention mechanism to assign higher scores to the frames in the clip that are more related to the narrated class.
Meanwhile, we extract multimodal video features from RGB frames, motion flow, and ambient sound. We apply a simple early fusion architecture to the problem and ablate the contributions of each modality.

To summarize, our contributions are as follows:
\begin{itemize}
    
    \item We propose to use the audio narrations to learn the video action detection model. To the best of our knowledge, this is a brand new task that has not been explored. In EPIC Kitchen tasks C1-C5, only C1-weakly is marginally related. However, C1-weakly is still different from ours because C1-weakly requires only to classify trimmed video at test time, while ours requires to localize actions in untrimmed videos.
    
    \item We provide a solution to the proposed task, in which we use a class-aware attention mechanism to rule out video frames that are not related to the narration label. Also, our solution considers multimodal features, including RGB frames, motion flow, and audio.
    
    \item We ablate our method on the EPIC Kitchens dataset and analyze the contributions of each model component and feature modality. The experiments provide insights for weakly supervised action detection methods in noisy untrimmed videos.
\end{itemize}

\section{Related Work}
\label{sec:related}

We note that there are many works \cite{Gkioxari_2015_CVPR,Ma_2016_CVPR,DBLP:conf/bmvc/YeBK18,Kazakos_2019_ICCV} focused on multimodal features in fully-supervised classification/detection tasks. In comparison, we use multimodal data as supervised signals to reduce human labeling efforts. We only summarize in this section the studies using multimodal signals to supervise the training process.

\textbf{Weakly Supervised Learning in Images.}
%We are aware that 
There is a large number of works aiming to harvest image models using weak supervision. Some use image-level labels \cite{Bilen_2016_CVPR,kantorov2016contextlocnet,Tang_2017_CVPR, tang2018pcl,Diba_2017_CVPR,Wei_2018_ECCV,Wan_2018_CVPR} or captions \cite{Ye_2019_ICCV,Zareian_2021_CVPR,chen2021towards} to learn object detection models. There are also works that learn semantic segmentation models \cite{tian2020cap2seg,vilar2021extracting,huynh2022open} or scene graph generation models \cite{Zhong_2021_ICCV,Ye_2021_CVPR}.
However, weakly supervised video models differ from all these prior image modeling approaches in that videos have the synchronized audio tracks, closed captions, and other information to be matched to the visual frames. Utilizing the additional synchronized modalities can potentially improve video models and is an important direction of learning video models using weak supervision.

\textbf{Weakly Supervised Learning in Videos.}
Since videos naturally involve multiple modalities, many approaches use unsupervised or weakly supervised training to learn better video representations. For example, \cite{aytar2016soundnet,Castrejon_2016_CVPR,ngiam2011multimodal,Owens_2016_CVPR} explore the cross-modal relations and leverage large amounts of unlabeled video for training. The basic idea behind this is that vision and sound are naturally synchronized so that models can utilize the synchronization as weak signals instead of ground truth labels. However, these methods are more often used in pre-training to improve the initialization of the visual-sound models. In comparison, our focus is on using additional modalities (e.g., audio narrations) to localize visual objects or actions in the temporal domain.

Also related is the co-localization or audio-visual correspondence~\cite{afouras2020self,Arandjelovic_2017_ICCV,Arandjelovic_2018_ECCV,Gao_2018_ECCV,Harwath_2018_ECCV,Senocak_2018_CVPR}. Similar to learning the joint representations, these works also rely on the synchronization of different modalities. However, they further learn to localize the sounds or visual objects given the information from other modalities. Our work differs from them in that (1) these works did not quantify their results on detection tasks while only providing qualitative results; (2) our model requires no supervised signals at testing time.

\textbf{Weakly Supervised Video Detection Tasks.}
In videos, there are various tasks of weakly supervised detection. For example, \cite{Singh_2017_ICCV,lee2020cross,Li_2021_CVPR,Nguyen_2018_CVPR,Mithun_2019_CVPR,Moltisanti_2019_CVPR,Wang_2017_CVPR} only learn to detect the starting and ending time of particular actions, while entirely ignoring the spatial layouts of the instances. \cite{DBLP:conf/bmvc/YeBK18} use audio amplitude from the video to predict the occurrence of ``climax'' in an advertisement video. To track the spatio-temporal localization, methods such as \cite{Chen_2017_CVPR,Weinzaepfel_2015_ICCV} rely on the video/image proposal frameworks such as \cite{Kang_2016_CVPR,kang2017t,Jain_2014_CVPR,uijlings2013selective} which provide high-quality region proposals. Their approaches are counterparts to weakly supervised object detection in the image domain, with the key difference being in the types of proposals. Finally, there are also methods~\cite{Misra_2015_CVPR,Singh_2016_CVPR,Kwak_2015_ICCV} attempting to only utilize cues from videos (e.g., motion, subtitle, tight boxes) to potentially benefit the training of image detectors.

We study how to learn action detection models (predicting starting/ending time and action labels) in videos. However, compared to fully-supervised methods \cite{Feichtenhofer_2019_ICCV,Gkioxari_2015_CVPR,Ma_2016_CVPR,Shou_2016_CVPR,Soomro_2015_ICCV,Yeung_2016_CVPR,Zhao_2017_ICCV}, the supervised signals we used are the audio narrations, which are noisy in nature hence are much weaker than instance-level annotations. As for the weakly supervised action detection models \cite{Singh_2017_ICCV,Nguyen_2018_CVPR,Nguyen_2019_ICCV,Paul_2018_ECCV,Shou_2018_ECCV,Wang_2017_CVPR}, their data in most cases only involves a single action per video. Thus video-level supervision satisfies their requirements. In comparison, our target task is a novel task, requiring non-trivial efforts to deal with the noisy annotations to improve the model's quality.

\section{Approach}
\label{sec:approach}

\begin{figure}[t]
    \centering
    \includegraphics[width=1\linewidth]{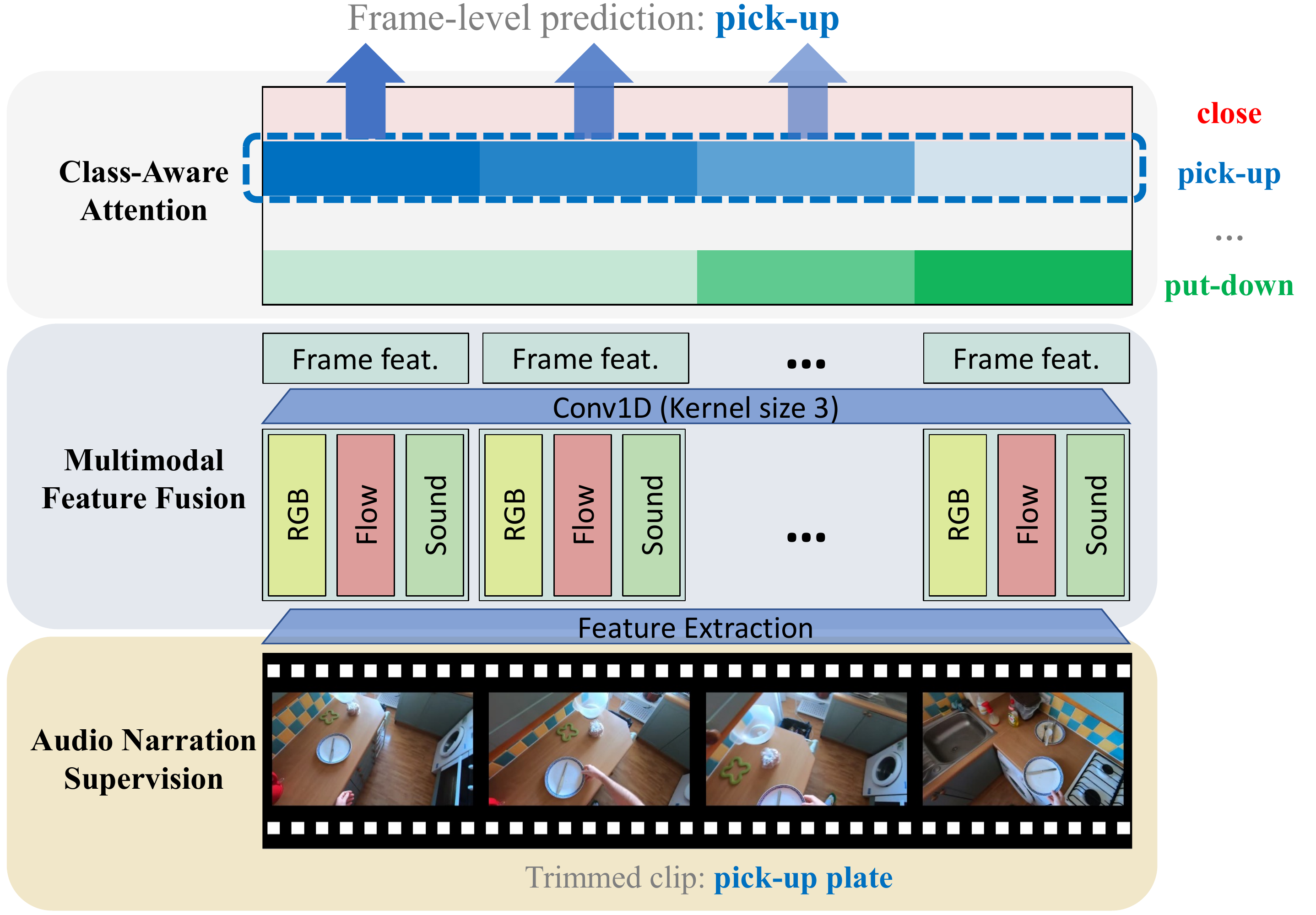}
    \caption[Model Overview]{Model overview. We first cut the video into clips using the single timestamp denoted in the audio narration. Then, each video clip can be treated as a bag of a few actions. Next, we extract multimodal features and use early fusion to combine them (Sec.~\ref{sec:approach:features}). We use a class-aware attention mechanism to produce the frame-level detection score (Sec.~\ref{sec:approach:attention}). Finally, we use a class-aware, intensity-sensitive post-processing (Sec.~\ref{sec:approach:postprocess}) to turn the frame-level into instance-level prediction (not shown), for evaluation purposes. }
    \label{fig:overview}
\end{figure}

We first formulate the weakly supervised action detection (WSAD) task guided by audio narration, and overview the model training pipeline. Then, we introduce the details regarding the multimodal features in Sec.~\ref{sec:approach:features}, discuss the design of the proposed class-aware attention in Sec.~\ref{sec:approach:attention}, and provide a post-processing algorithm which turns the frame-level into instance-level prediction (required by evaluation), in Sec.~\ref{sec:approach:postprocess}.

\textbf{Task formulation:} We conduct our experiments on the EPIC Kitchens dataset~\cite{Damen_2018_ECCV}. At training time, the video and paired $\{time_i, verb_i, noun_i\}_{i=1}^{N}$ as $N$ annotated actions are provided, where $time_i$ is the narration start time, and $verb_i$ and $noun_i$ are the narrated verb and noun classes respectively. The underlying assumption is that $time_i$ is not precise to represent the narration starting time since there may be overlap between consecutive actions. At test time, models have to predict four tuples  $\{time\_s_i, time\_e_i, verb_i, noun_i\}$ given the video, where $time\_s_i, time\_e_i$ are the start and end time respectively.

\textbf{Training pipeline overview (Fig.~\ref{fig:overview}):} Given a video and the paired audio narration annotation $\{time_i, verb_i, noun_i\}_{i=1}^{N}$, we first split the video into training clips. Given a specific action $(time_i, verb_i, noun_i), i\in\{1\dots N\}$, we cut the video from $time_i$ to $time_{i+1}$, resulting in a video clip ($N$ clips in total) paired with $verb_i$ and $noun_i$. We denote the frames in the $i$-th clip as $\{f_{i,j}\}_{j=1}^{L_i}$ where $L_i$ is the total number of frames in the $i$-th video clip.

Then (Fig.~\ref{fig:overview} (middle)), we proceed with the feature extraction process, which will be explained in detail in Sec.~\ref{sec:approach:features}. Briefly, we extract the visual CNN features of the RGB and flow frames and the semantic embedding of the ambient soundtrack. After feature extraction, we use early fusion to aggregate these multiple modalities.
% Next, we interpolate all these features to be 1 frame per second for feature fusion since they may differ in sampling rate.

Finally, we use an attention mechanism guided by audio narrations, 
%classes, 
to filter out irrelevant classes in the clip, given that the $i$-th clip should be all regarding the $verb_i$ and $noun_i$. 
In Fig.~\ref{fig:overview} (top), we show an example clip related to the ``pick-up plate'' action, and the figure demonstrates how to predict the verb ``pick-up''. Because of \textbf{the nature of the noisy narration supervision}, it is common that the last few frames in the clip mix with the next action (e.g., ``put-down plate''). So given the clip level supervision ``pick-up'', how can we rule out the impacts of the next action ``put-down''? Our solution is to predict different attention distributions for different verbs (in Fig.~\ref{fig:overview} (top), blue for ``pick-up'' and ``green'' for ``put-down'').
%attention distributions). 
Then, within a clip, we use the attention distribution bound to the action class (e.g., verb ``pick-up'' is the supervision for the clip in Fig.~\ref{fig:overview} (top)) to help with the predictions. The selected attention distribution will highlight the frames for predicting the clip-level action.
% For example, in Fig.~\ref{fig:overview} (top), we show that given the verb class ``put-down'' as the query, the attention will focus on the last few frames (potentially have overlap with the following action). 
% However, given the narration supervision of the clip, we need to choose the attention distribution of the verb ``pick-up'' among frames. The frames highlighted by the attention scores are then responsible for predicting the clip-level action (e.g., ``pick-up'').

\subsection{Multimodal Video Features}
\label{sec:approach:features}

We consider features from the following sources. Though using them is common in video recognition, we are working on a novel task of learning action detection models from narration supervision, in which the contributions of RGB/flow/audio features are unclear.

\begin{itemize}
    \item \textbf{RGB and flow frames.} We use the standard RGB and flow features provided in the EPIC Kitchens dataset, i.e., the 1024-D RGB and 1024-D Flow CNN features generated by a TSN model\cite{Furnari_2019_ICCV} pre-trained on \cite{Damen_2018_ECCV}.
    
    \item \textbf{Ambient sound.} Since the EPIC Kitchens dataset provides the soundtrack of the ambient audio, we also model them because the sound may imply some actions. We use  VGGish~\cite{gemmeke2017audio} to produce a 128-D semantically meaningful embedding for every second. The VGGish method was first used in the AudioSet~\cite{gemmeke2017audio} classification task, and it was pre-trained on a large YouTube dataset (which later became YouTube-8M).

\end{itemize}

\textbf{Early fusion of the multimodal video features.}
We linearly interpolate the ambient sound semantic embeddings to convert its sequence lengths to be the same as the RGB and flow features. 
We denote the concatenation of these multimodal features as the video frame feature $\{\bm{f}_{i, j}\}_{j=1}^{L_i}$, $\bm{f}_{i,j} \in \mathbb{R}^{2176 \times 1}$ (RGB 1024-D, flow 1024-D, ambient sound 128-D, $L_i$ - number of frames).
Let $\bm{F}_i = [\bm{f}_{i,1}~\bm{f}_{i,2} \cdots \bm{f}_{i,L_i}]^T \in \mathbb{R}^{L_i \times 2176}$ be the video sequence feature, we apply a Conv1D layer (with kernel size 3, ReLu activation) to further extract the frame feature $\mathbf{F}_i = [\mathbf{f}_{i, 1}~\mathbf{f}_{i, 2} \cdots \mathbf{f}_{i, L_i}]^T \in \mathbb{R}^{L_i \times d}$ ($d=100$ is the number of neuron units):

\begin{equation} \label{epic:eq:conv1d}
    \mathbf{F}_i = \text{Conv1D} (\bm{F}_i)
\end{equation}

\subsection{Class-Aware Attention for Weakly Supervised Action Detection}
\label{sec:approach:attention}

After getting the multimodal video features, we design a class-aware attention mechanism to localize the actions in the sequences. Our model selects relevant frames that best represent the action in the video clip and uses their aggregated features to represent it.
Take Fig.~\ref{fig:overview} as an example, the verb class for the video clip is ``pick-up'' (i.e., clip-level label), so we use the embedding of ``pick-up'' to multiply (dot-product) each frame feature $\mathbf{f}_i$ to measure the frame-label similarity, resulting in a sequence of scores. After normalization, this score array represents the likelihood that the associated frames involve the action ``pick-up''. We compute the weighted sum of the sequence features (weighed by the normalized scores). Then, we add a classification layer to predict the action and use cross-entropy loss to optimize.

Formally, we define action label embedding weights $\mathbf{W}_{verb}^{(1)} \in \mathbb{R}^{C_{verb} \times d}$, $\mathbf{W}_{noun}^{(1)} \in \mathbb{R}^{C_{noun}\times d}$ where $C_{verb}$ and $C_{noun}$ are the number of verb and noun classes, respectively. Since the verb detection and noun detection follow the same pipeline and only differ in the number of classes, we use $\mathbf{W}^{(1)} = \mathbf{W}_{verb}^{(1)}$ or $\mathbf{W}_{noun}^{(1)}$ as an abstract notation to denote either label embedding, $C = C_{verb}$ or $C_{noun}$ to denote the number of classes, and $c_i=verb_i$ or $noun_i$ to denote the clip-level label. Then, the following procedure applies to both verb and noun detection parallel tasks.

We first compute the dot-product between the label embedding and the frame feature, then, we use the sigmoid function to turn the score into a probability $\mathbf{A}_i' \in \mathbb{R}^{C \times L_i}$ (see Eq.~\ref{epic:eq:attention}; Fig.~\ref{fig:overview} (top) shows $\mathbf{A}_i'$ using the color matrix).
Since we are aware of the class that is narrated in the video clip, we select the specific $c_i$-th row in $\mathbf{A}_i'$ ($c_i = verb_i$ or $noun_i$), resulting in $\mathbf{A}_i \in \mathbb{R}^{1\times L_i}$. This class-aware row selection process is shown in Fig.~\ref{fig:overview} (top) using the blue dashed box.

\begin{equation} \label{epic:eq:attention}
    \mathbf{A}_i' = \text{sigmoid}(\mathbf{W}^{(1)} ~ \mathbf{F}_i^T), \quad\quad
    \mathbf{A}_i = \mathbf{A}_i'[c, :]
\end{equation}

Meanwhile, we use a fully connected layer $\mathbf{W}^{(2)} \in \mathbb{R}^{d\times C}$ to estimate the per-frame detection score $\mathbf{D}_i \in \mathbb{R} ^{L_i \times C}$. In Eq.~\ref{epic:eq:detection}, $j \in \{1 \cdots L_i\}$ denotes frame id and $k \in \{1 \cdots C\}$ is the class index.

\begin{equation} \label{epic:eq:detection}
    \mathbf{D}_i' = \mathbf{F}_i ~ \mathbf{W}^{(2)},  \quad\quad 
    \mathbf{D}_i[j,k] = \frac{\exp{(\mathbf{D}_i'[j,k])}}{\sum_{k'=1}^{C}\exp{(\mathbf{D}_i'[j,k'])}}
\end{equation}

Directly optimizing the per-frame detection score $\mathbf{D}_i = \mathbf{D}_{verb~i}$ or $\mathbf{D}_{noun~i}$ is hard since we only have the clip-level label $c_i = verb_i$ or $noun_i$. Thus, we apply the class-aware attention weighting $\mathbf{A}_i$ to aggregate frame-level information into $\bar{\mathbf{F}_i}\in \mathbb{R}^{1\times d}$ (Eq.~\ref{epic:eq:weighting}), which is a clip-level feature.
Then, the clip-level prediction is given by $\mathbf{P}_i \in \mathbb{R}^{C\times 1}$ (Eq.~\ref{epic:eq:prediction}), which shares the $\mathbf{W}^{(2)}$ with Eq.~\ref{epic:eq:detection}.

\begin{equation} \label{epic:eq:weighting}
    \bar{\mathbf{F}_i}=\frac{\mathbf{A}_i \mathbf{F}_i}{\sum_{j=1}^{L_i}\mathbf{A}_i[j]}
\end{equation}

\begin{equation} \label{epic:eq:prediction}
    \mathbf{P}_i'=(\bar{\mathbf{F}_i} ~ \mathbf{W}^{(2)})^T, \quad \quad
    \mathbf{P}_i[k] = \frac{\exp{(\mathbf{P}_i'[k])}}{\sum_{k'=1}^{C} \exp{(\mathbf{P}_i'[k'])}}
\end{equation}

Finally, we use cross-entropy to optimize the model, where $\bm{y}_i$ is the one-hot representation of $c_i$ ($\bm{y}_i[k]=1 \text{ iff } k=c_i$).
\begin{equation} \label{epic:eq:loss}
    L = - \sum_i \sum_{k=1}^{C} \bm{y}_i[k]\log{\mathbf{P}_i[k]}
\end{equation}

\subsection{Class-Aware Intensity-Sensitive Post Processing}
\label{sec:approach:postprocess}

\begin{figure}[t]
    \centering
    \includegraphics[width=1.0\linewidth]{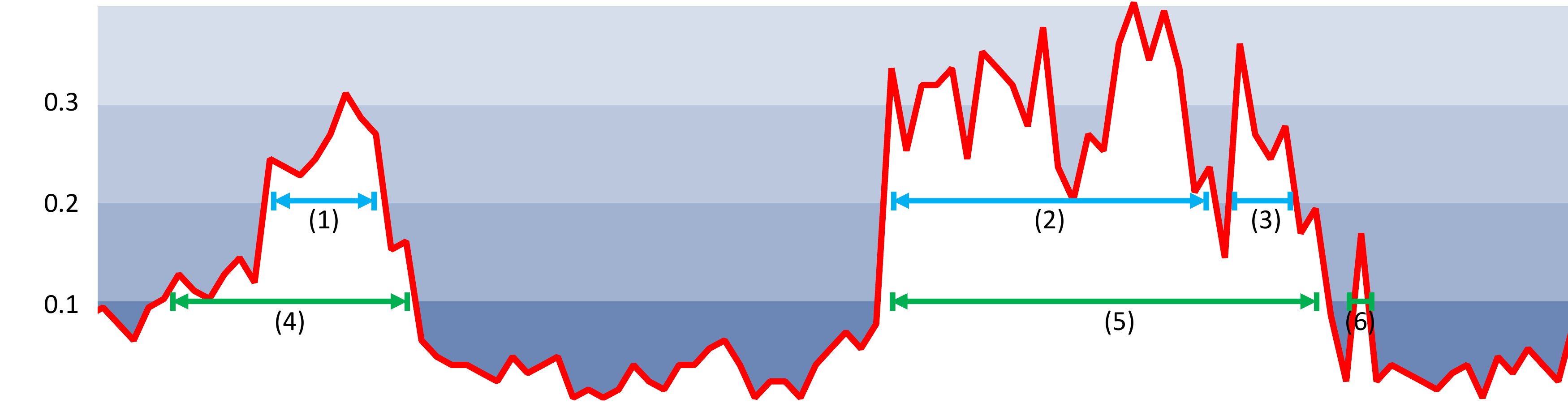}
    \caption[Intensity-Sensitive Post-Processing]{Intensity-sensitive post-processing. For each of the action classes, we use a set of thresholds (e.g., \{0.1, 0.2\}) and retrieve all segments (consecutive frames) that meet the different threshold conditions. The retrieval results are a set of action segments with different intensities.
    Next, we score each segment and apply Non-Maximum Suppression (NMS) to remove highly overlapped detections. We show the action clips detected using a threshold of 0.1 using green color and the clips detected by threshold 0.2 using blue. Assuming the IoU threshold of 0.6, segment (5) will be removed because it  overlapped with (2).}
    \label{fig:postprocess}
\end{figure}

To get the detections in the form of $\{time\_s_i, time\_e_i, verb_i, noun_i\}$ from the frame-level prediction $\mathbf{D}_i$ (Eq.~\ref{epic:eq:detection}), we use a class-aware intensity-sensitive post process. Specifically, we consider each action class separately.
Given the detection score of a specific class (e.g., the $k$-th class in verb detection $\mathbf{D}_{verb~i}[:, k]$), we first use different intensities (thresholds) to retrieve the segments, which are defined to be the longest sequence of consecutive frames that have detection scores past the threshold. The result is a set of potential action segments detected by different intensity scores (thresholds).
We then assign a score to each segment, denoting the average detection intensity within the segment. In Fig.~\ref{fig:postprocess}, we show the segments detected by threshold 0.1 and 0.2 using green and blue colors, respectively.
In the next step, we use Non-Maximum Suppression (NMS) to remove highly overlapped (measured by IoU) detections and retain only those with higher intensity.
Finally, we aggregate the NMS-ed detections from all action classes and sort them by intensity, resulting in our final detection results.

\section{Experiments}
\label{sec:results}

We provide the details regarding our model in Sec.~\ref{sec:results:details}. Then, we provide experimental results in Sec.~\ref{sec:results:epic}, including analysis regarding both the contributions of our model components and the benefits of multimodal features.
To better understand our model, we also provide qualitative results in Sec.~\ref{sec:results:qualitative}.

\subsection{Implementation Details}
\label{sec:results:details}

Before training the detector, we extract the multimodal features offline. The CNN features of the RGB and flow frames are from \cite{Damen_2018_ECCV}, while we pre-processed the audio features. We use FFMpeg to extract audios from MP4 videos and feed the Mel spectrogram to the VGGish~\cite{gemmeke2017audio} model pre-trained on the large Youtube dataset (latter Youtube-8M) to produce semantic audio embeddings.
After getting the above features, we interpolate the audio features to make them the same lengths as the RGB and flow features.

\begin{table*}[t]
    \centering
    \setlength{\tabcolsep}{1pt}
    \begin{tabularx}{1\linewidth}{c|*{6}{>{\centering\arraybackslash}X}|*{6}{>{\centering\arraybackslash}X}|*{6}{>{\centering\arraybackslash}X}}
    \Xhline{2\arrayrulewidth}
        \multirow{2}{*}{} & \multicolumn{6}{c|}{Action Detection} & \multicolumn{6}{c|}{Verb Detection} & \multicolumn{6}{c}{Noun Detection} \\
        & @0.1 & @0.2 & @0.3 & @0.4 & @0.5 & Avg. & @0.1 & @0.2 & @0.3 & @0.4 & @0.5 & Avg. & @0.1 & @0.2 & @0.3 & @0.4 & @0.5 & Avg. \\
    \Xhline{1\arrayrulewidth}
        \textsc{Ful.~\cite{damen2020rescaling}}   & 6.95 & 6.10 & 5.22 & 4.36 & 3.43 & 5.21 & 10.8 & 9.84  & 8.43 & 7.11 & 5.58 & 8.36 & 10.3 & 8.33 & 6.17 & 4.47 & 3.35 & 6.53 \\
        \textsc{Our Ful.}                         & 6.40 & 5.69 & 4.59 & 3.34 & 2.39 & 4.48 & 12.9 & 11.4 & 9.04 & 6.62 & 5.03 & 9.00 & 11.4 & 9.61 & 7.17 & 4.70 & 2.98 & 7.17 \\
    \Xhline{1\arrayrulewidth}
        \textsc{Narr. Bas.} & 4.42 & 3.62  & 2.91 & 2.06 & 1.47 & 2.90 & 9.39  & 7.45  & 5.68 & 3.99 & 2.85 & 5.87 & 8.43  & \textbf{6.92} & \textbf{5.24} & \textbf{3.50} & \textbf{2.37} & \textbf{5.29} \\
        \textsc{Cls. Agno.} & \textit{4.57} & \textit{3.78} & \textit{3.10} & \textit{2.28} & \textbf{1.70} & \textit{3.09} & \textbf{10.0} & \textbf{8.53}  & \textbf{7.03} & \textbf{4.79} & \textit{3.40} & \textbf{6.75} & \textit{8.49}  & 6.82 & 4.96 & 3.22 & 2.04 & 5.11 \\
        Ours                     & \textbf{4.68} & \textbf{4.01} & \textbf{3.27} & \textbf{2.33} & \textit{1.65} & \textbf{3.19} & \textit{9.64}  & \textit{7.96}  & \textit{6.31} & \textit{4.70} & \textbf{3.56} & \textit{6.43} & \textbf{8.51} & \textit{6.88} & \textit{5.09} & \textit{3.36} & \textit{2.25} & \textit{5.22} \\
    \Xhline{2\arrayrulewidth}
    \end{tabularx}
    \caption[Contributions of Proposed Components]{Contributions of proposed components. We show the Average Precision (\%) at certain IoU thresholds (@0.1-@0.5) and the mean Average Precision (Avg.). Higher numbers are better. The best weakly supervised model learned using narration annotations is shown in \textbf{bold} and the second best is in \textit{italic}.}
    \label{tab:alternatives}
\end{table*}
\begin{table*}[t]
    \centering
    \setlength{\tabcolsep}{1pt}
    \begin{tabularx}{1\linewidth}{c|*{6}{>{\centering\arraybackslash}X}|*{6}{>{\centering\arraybackslash}X}|*{6}{>{\centering\arraybackslash}X}}
    \Xhline{2\arrayrulewidth}
        \multirow{2}{*}{} & \multicolumn{6}{c|}{Action Detection} & \multicolumn{6}{c|}{Verb Detection} & \multicolumn{6}{c}{Noun Detection} \\
        & @0.1 & @0.2 & @0.3 & @0.4 & @0.5 & Avg. & @0.1 & @0.2 & @0.3 & @0.4 & @0.5 & Avg. & @0.1 & @0.2 & @0.3 & @0.4 & @0.5 & Avg. \\
    \Xhline{1\arrayrulewidth}
        RGB    & 4.49 & 3.76 & 2.94 & 2.25 & \textbf{1.67} & 3.02 & 8.72 & 7.07 & 5.43 & 4.38 & 3.15 & 5.75 & \textbf{8.70} & \textbf{7.13} & \textbf{5.29} & \textbf{3.71} & \textbf{2.56} & \textbf{5.48} \\
        Flow   & 2.32 & 1.98 & 1.47 & 1.10 & 0.84 & 1.54 & 6.59 & 5.58 & 4.29 & 2.95 & 2.10 & 4.30 & 4.33 & 3.47 & 2.49 & 1.68 & 1.11 & 2.61 \\
        Audio  & 0.34 & 0.27 & 0.23 & 0.09 & 0.05 & 0.20 & 1.71 & 1.37 & 1.07 & 0.61 & 0.39 & 1.03 & 0.94 & 0.68 & 0.51 & 0.23 & 0.16 & 0.50 \\
    % \Xhline{1\arrayrulewidth}
    %     wo/ RGB   & 2.48 & 2.13 & 1.66 & 1.28 & 0.98 & 1.70 & 7.06 & 6.03 & 4.79 & 3.56 & 2.86 & 4.86 & 4.49 & 3.51 & 2.57 & 1.74 & 1.13 & 2.69 \\
    %     wo/ Flow  & 4.39 & 3.68 & 2.90 & 2.11 & 1.53 & 2.92 & 8.66 & 7.14 & 5.49 & 4.01 & 3.11 & 5.68 & 8.92 & 7.17 & 5.26 & 3.49 & 2.31 & 5.43 \\
    %     wo/ Audio & 4.82 & 4.14 & 3.34 & 2.53 & 1.88 & 3.34 & 9.78 & 8.46 & 6.90 & 4.59 & 3.51 & 6.65 & 8.62 & 6.91 & 5.01 & 3.47 & 2.16 & 5.23 \\
    \Xhline{1\arrayrulewidth}
        All    & \textbf{4.68} & \textbf{4.01} & \textbf{3.27} & \textbf{2.33} & 1.65 & \textbf{3.19} & \textbf{9.64} & \textbf{7.96} & \textbf{6.31} & \textbf{4.70} & \textbf{3.56} & \textbf{6.43} & 8.51 & 6.88 & 5.09 & 3.36 & 2.25 & 5.22 \\
    \Xhline{2\arrayrulewidth}
    \end{tabularx}
    \caption[Contributions of Multimodal Features]{Contributions of multimodal features. We show the Average Precision (\%) at certain IoU thresholds (@0.1-@0.5) and the mean Average Precision (Avg.). Higher numbers are better. The best model is shown in \textbf{bold}.}
    \label{tab:features}
\end{table*}

We concatenate the multimodal features as the model input and add a Conv1D layer (with $d=100$ filters, kernel size 3, ReLu activation) to further finetune. During training, we use a dropout probability of 0.5 for the Conv1D layer, and a dropout probability of 0.5 for the learned attention ($\mathbf{A}_i$). We use the Tensorflow framework~\cite{abadi2016tensorflow}, Adam optimizer~\cite{kingma2014adam}, a learning rate of 1e-5, and a batch size of 8 (8 clips). All models in our experimental sections are trained for 300K steps on the EPIC Kitchens dataset, using a validation set to pick the best model.

For the post-processing, we first apply uniform filtering (filter size 3) on each class's detection scores (e.g., $\mathbf{D}_{verb~i}[:, k]$) to make the detection scores less fluctuating. Then, we vary the detection threshold from 0.01 to 0.4 to retrieve all segments and use NMS with an IoU threshold of 0.4 to remove highly overlapped segments.

\subsection{Results on the EPIC Kitchens Dataset}
\label{sec:results:epic}

\textbf{Metrics.} 
Although our training process does not rely on instance-level annotations, we can use the EPIC Kitchens' C2 task's (Action Detection) evaluation protocol, which measures the performance of the action detections. The protocol computes the average of the Average Precision (AP) values for each class, a.k.a. mean AP. A predicted segment is considered correct if its Intersection over Union (IoU) with a ground truth segment is greater than or equal to a given threshold (0.1 to 0.5).
Besides the verb and noun detection, the EPIC Kitchens' C2 task also involves an action detection evaluation which requires the verb and noun detections to be correct at the same time.

\textbf{Contributions of Proposed Components.}
We verify the effectiveness of the proposed model and compare it to the fully- and weakly-supervised action detection methods. All methods listed below use the same features.

\begin{itemize}
    \item \textsc{Ful.~\cite{damen2020rescaling}} is a fully supervised model trained by the EPIC Kitchens challenge organizer, using a two-stage approach to solve the action detection (action proposal~\cite{Lin_2019_ICCV} + action classification~\cite{Feichtenhofer_2019_ICCV}). 
    
    \item \textsc{Our Ful.} is a one-stage fully supervised method trained by us, in which we predict the frame-level actions then post-process (Sec.~\ref{sec:approach:postprocess}). We treat \textsc{Our Ful.} as a proper upper bound baseline in that all of our weakly supervised methods depend on similar frame-level prediction + post-processing.

    \item \textsc{Narr. Bas.} is the baseline method of using narration supervision. In \textsc{Narr. Bas.}, we treat the single timestamp in the narration annotations as the boundaries and use the cut result as instance-level annotations to directly train a fully supervised model.

    \item \textsc{Cls. Agno.} is an alternative method, in which we use a class-agnostic attention instead of class-aware attention (Sec.~\ref{sec:approach:attention}).
\end{itemize}

\begin{table*}[t]
    \centering
    \setlength{\tabcolsep}{0pt}
    \begin{tabularx}{1\linewidth}{*{6}{>{\centering\arraybackslash}X}|*{6}{>{\centering\arraybackslash}X}}
    \Xhline{2\arrayrulewidth}
        \multicolumn{6}{c|}{Verb Detection} & \multicolumn{6}{c}{Noun Detection} \\
        \multicolumn{2}{c}{RGB} & \multicolumn{2}{c}{Flow} & \multicolumn{2}{c|}{Audio} & \multicolumn{2}{c}{RGB} & \multicolumn{2}{c}{Flow} & \multicolumn{2}{c}{Audio} \\   
        Name & AP(\%) & Name & AP(\%) & Name & AP(\%) & Name & AP(\%) & Name & AP(\%) & Name & AP(\%) \\
    \Xhline{1\arrayrulewidth}
        wash   & 39.59 & wash  & 37.14 & wash    & 17.47 & corn    & 33.21 & yoghurt & 28.50 & \scriptsize{microwave} & 18.33 \\
        filter & 31.33 & hang  & 29.69 & season  & 11.81 & raisin  & 33.17 & tray    & 21.81 & salt      & 11.79 \\
        rip    & 30.55 & fold  & 23.32 & measure & 8.12  & yoghurt & 33.17 & lid     & 21.14 & oatmeal   & 5.63  \\
        season & 30.11 & dry   & 19.88 & unscrew & 6.65  & olive   & 29.34 & cloth   & 20.72 & carrot    & 5.27  \\
        fold   & 25.51 & throw & 17.75 & squeeze & 5.11  & lid     & 28.16 & oven    & 18.67 & \scriptsize{cupboard}  & 4.31  \\
    \Xhline{2\arrayrulewidth}
    \end{tabularx}
    \caption[Classes Detected by the Multimodal Features]{Top-5 classes detected by the RGB, Flow, and Audio features.}
    \label{tab:detected}
\end{table*}
\begin{figure*}[t]
    \centering
    \includegraphics[width=0.7\linewidth]{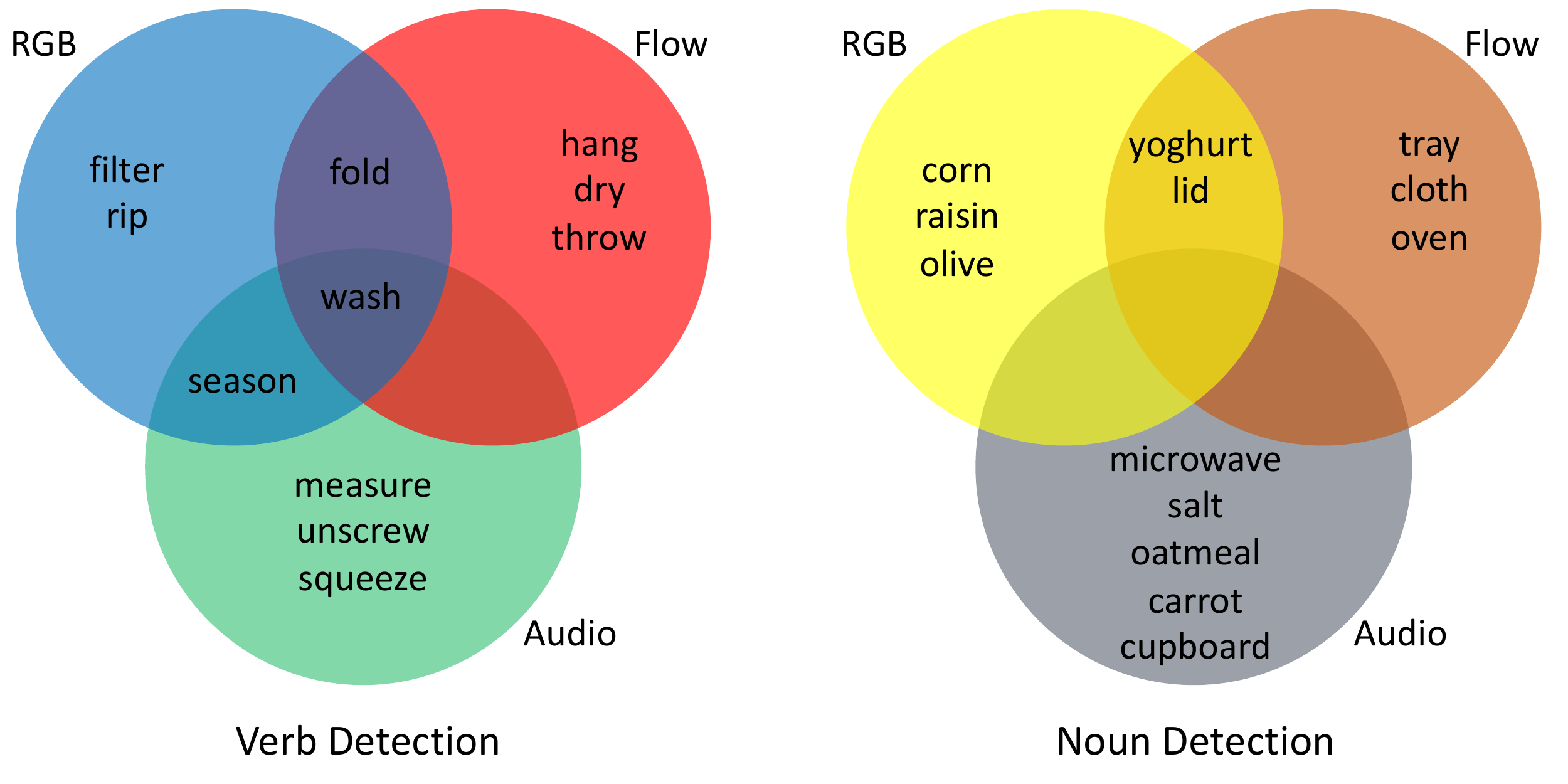}
    \caption[The top-5 Classes Detected by Different Modalities]{Venn diagrams - The easily detected top-5 classes by different modalities. }
    \label{fig:venn}
\end{figure*}

Tab.~\ref{tab:alternatives} shows the results.
We found \textsc{Our Ful.}, though a one-stage method, is very competitive to \textsc{Ful.~\cite{damen2020rescaling}} (action detection mAP 4.48\% v.s. 5.21\%). The only weakness is that it is not that good at boundary refinement. Hence its verb and noun detection AP@0.1,0.2,0.3 are higher, but its AP@0.4,0.5 are lower.
Then, \textsc{Narr. Bas.}, which uses the same fully supervised method (but changes to use the narration supervision), inevitably hurts the action detection performance (action mAP 2.90\% v.s. 4.48\%).
This performance drop is due to the unclear boundary definition. We conclude that our method with uncertainty modeling (class-aware attention) helps to improve the use of narration supervision (action mAP 3.19\% v.s. 2.90\%). Also, we show that our modeling of class-aware attention is better than the alternative of class-agnostic attention (action mAP 3.19\% v.s. 3.09\%). The reason, we argue, is that the class-agnostic attention is only able to distinguish the dynamic actions from the background frames (e.g., solving the task in THUMOS 14 as shown in Tab.~\ref{tab:datasets}). It fails if the mixed actions are all semantically meaningful video frames.

\textbf{Contributions of Multimodal Features.}
We analyze the contributions of multimodal features by building our models on different subsets of features. We first build our models using single modalities, then present our model considering all types of features. 

Tab.~\ref{tab:features} shows the results.
Among the single modality models, the RGB model provides the best performance on Action Detection (mAP 3.02\%). It achieves both high verb detection (mAP 5.75\%) and high noun detection (mAP 5.48\%) performance. The flow model (action mAP 1.54\%) is worse than the RGB model but is better than the Audio model. We can see that the flow feature provides more information for the dynamic actions (verb mAP 4.30\%), while it is not that good at localizing static objects temporally (noun mAP 2.61\%). The audio model (action mAP 0.20\%) is the worst among the three single modal models, but it still provides useful information, especially in verb detection (mAP 1.03\%).

Our final model takes advantage of all features and achieves the best performance in terms of action detection mAP (3.19\%). Compared to the RGB model, it utilizes the flow and audio information to better detect the dynamic actions (verb mAP 6.43\% v.s. 5.75\%). Furthermore, compared to the flow and audio models, it combines the appearance feature (RGB) to better recognize objects in the temporal domain (noun mAP 5.22\% v.s. 2.61\%, 0.50\%).
In sum, we conclude that our modeling of the videos' multimodal nature helps improve the weakly supervised action detection task.

We show in Tab.~\ref{tab:detected} and Fig.~\ref{fig:venn} the verb and noun classes best detected by the three modalities.
For the verb detection (Fig.~\ref{fig:venn} (left)), action ``wash'' can be easily detected by all three modalities, while ``fold'' only makes a slight sound, so it is hard to recognize by audio. In comparison, ``season'' sounds loud, but the dynamic action is nuanced; thus, the audio can detect it but not the motion flow.
The noun detection results are also interesting (Fig.~\ref{fig:venn} (right)). We found the ``tray'' and ``cloth'' to be more dynamic, and we notice that ``microwave'' makes a sound. So, we conclude that different modalities help localize different objects and actions temporally.

We are aware that a similar Venn gram is provided in \cite{Kazakos_2019_ICCV}, but our findings are different because of the task difference. Besides using ambient sound, we emphasize that we used audio narration supervision to reduce the human labeling efforts, which had not been explored in \cite{Kazakos_2019_ICCV}.

\subsection{Qualitative Examples}
\label{sec:results:qualitative}

\begin{figure*}[t]
    \centering
    \includegraphics[width=1.0\linewidth]{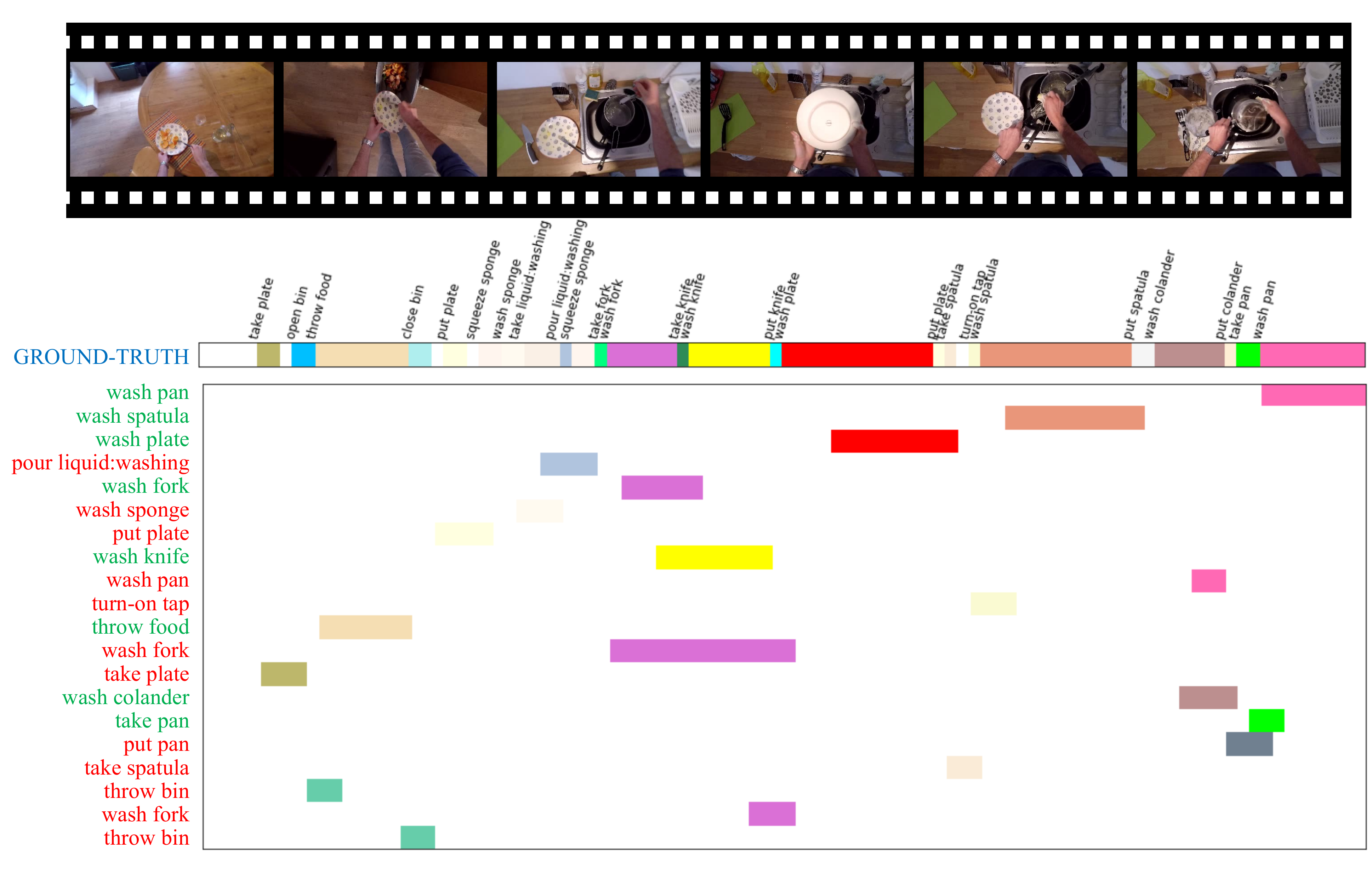}
    \caption[Qualitative Example of Our Model's Action Detection Results]{Qualitative example of our model's action detection results. We show the demo of the video, the ground-truth annotations, and our model's top-20 predictions. We show the correct predictions using green and incorrect ones using red. The correctness is determined by IoU@0.5.}
    \label{fig:qualitative}
\end{figure*}

We provide a qualitative example visualizing the results of our model. Fig.~\ref{fig:qualitative} shows it confidently and correctly localizes the actions ``wash pan'', ``wash spatula'', and ``wash plate''. For actions such as ``pour liquid:washing'' and ``wash sponge'', our model's estimations of the starting and ending time are not precise, thus causing the IoU with the ground truth to be smaller than 0.5. We can hardly find mistakes regarding classification issues in the top-20. Hence we conclude that localization and refining the action boundaries are still challenging for weakly supervised action detection and should gather more attention.

\section{Conclusion}
\label{sec:conclusion}

We explored audio narration as a form of supervision in this paper. We developed a model to learn from the narration supervision and utilize multimodal features, including RGB, motion flow, and ambient sound. In our design, the model learns to attend to the frames related to the narration label while suppressing the irrelevant frames from being used. In the experiments, we show that the proposed method outperformed alternative designs. Also, we proved that the different modalities contribute to the detections of different actions and objects in the temporal domain.

The insights of our paper are interesting. Throughout the EPIC Kitchen tasks C1-C5, none of them directly uses the cheap audio narration supervision to learn action detectors, while we proved such a task of using narration to be possible. Our experiments have shown that it is plausible to eliminate the expensive stages of refining action boundaries during video detection data annotation. Further, the refined instance-level annotations did not contribute too much to the detector's performance. We expect weakly-, semi-, and self-supervised methods to gather more attention in future video detection tasks.

\noindent \textbf{Acknowledgement:} This work was supported by a University of Pittsburgh CS50 Computer Science fellowship, and by National Science Foundation award 2046853.

%%%%%%%%% REFERENCES
{\small
\bibliographystyle{ieee_fullname}
\bibliography{egbib}
}

\end{document}